\documentclass[]{article}
\usepackage[letterpaper]{geometry}
\usepackage{mtsummit2021}
\usepackage{times}
\usepackage{url}
\usepackage{latexsym}
\usepackage{natbib}
\usepackage{layout}
\usepackage{graphicx}
\usepackage[normalem]{ulem}
\useunder{\uline}{\ul}{}

\usepackage{multicol}

\usepackage{appendix}
\usepackage{float}
\restylefloat{table}
\usepackage{amsmath}

\usepackage{caption}
\usepackage{subcaption}

\begin{document}

\title{\bf Machine Translation in the Covid domain: an English-Irish case study for LoResMT 2021}

\author{\name{\bf Séamus Lankford} \hfill  \addr{seamus.lankford@adaptcentre.ie}\\
        \addr{ADAPT Centre, School of Computing, Dublin City University, Dublin, Ireland.}
        \AND
       \name{\bf Haithem Afli} \hfill \addr{haithem.afli@adaptcentre.ie}\\
        \addr{ADAPT Centre, Department of Computer Science, Munster Technological University, Ireland.}
        \AND        
        \name{\bf Andy Way} \hfill \addr{andy.way@adaptcentre.ie}\\
        \addr{ADAPT Centre, School of Computing, Dublin City University, Dublin, Ireland.}
}
\maketitle
\pagestyle{empty}

\begin{abstract}

Translation models for the specific domain of translating Covid data from English to Irish were developed for the LoResMT 2021 shared task. Domain adaptation techniques, using a Covid-adapted generic 55k corpus from the Directorate General of Translation, were applied. Fine-tuning, mixed fine-tuning and combined dataset approaches were compared with models trained on an extended in-domain dataset.  As part of this study, an English-Irish dataset of Covid related data, from the Health and Education domains, was developed. The highest-performing model used a Transformer architecture trained with an extended in-domain Covid dataset. In the context of this study, we have demonstrated that extending an 8k in-domain baseline dataset by just 5k lines improved the BLEU score by 27 points. 

\end{abstract}

\section{Introduction}
Neural Machine Translation (NMT) has routinely outperformed Statistical Machine Translation (SMT) when large parallel datasets are available~\citep{crego2016systran, wu2016google}. Furthermore, Transformer based approaches have demonstrated impressive results in moderate low-resource scenarios~\citep{lankford2021Transformer}. 
NMT involving Transformer model development will improve the performance in specific domains of low-resource languages ~\citep{araabi2020optimizing}. However, the benefits of NMT are less clear when using very low-resource Machine Translation (MT) on in-domain datasets of less than 10k lines. 

The Irish language is a primary example of a low-resource language that will benefit from such research. This paper reports the results for the MT system developed for the English–Irish shared task at LoResMT 2021~\citep{ojha-etal-2021-findings}. Relevant work is presented in the background section followed by an overview of the proposed approach. The empirical findings are outlined in the results section. Finally, the key findings are presented and discussed. 

\section{Background}
   
\subsection{Transformer}

A novel architecture called Transformer was introduced in the paper ‘Attention Is All You Need’ ~\citep{vaswani2017attention}. Transformer is an architecture for transforming one sequence into another with the help of an Encoder and Decoder without relying on Recurrent Neural Networks.

Transformer models use attention to focus on previously generated tokens. This approach allows models to develop a long memory which is particularly useful in the  domain of language translation. 

\subsection{Domain adaptation}

Domain adaptation is a proven approach in addressing the paucity of data in low-resource settings. Fine-tuning an out-of-domain model by further training with in-domain data is effective in improving the performance of translation models (Freitag and Al-Onaizan, 2016; Sennrich et al., 2016). With this approach an NMT model is initially trained using a large out-of-domain corpus. Once fully converged, the out-of-domain model is further trained by fine-tuning its parameters with a low resource in-domain corpus. 

A modification to this approach is known as mixed fine-tuning~\citep{chu2017empirical}. With this technique, an NMT model is trained on out-of-domain data until fully converged. This serves as a base model which is further trained using the combined in-domain and out-of-domain datasets. 

\section{Proposed Approach}

\begin{figure} [H]
    \centering
    \includegraphics[width=14cm]{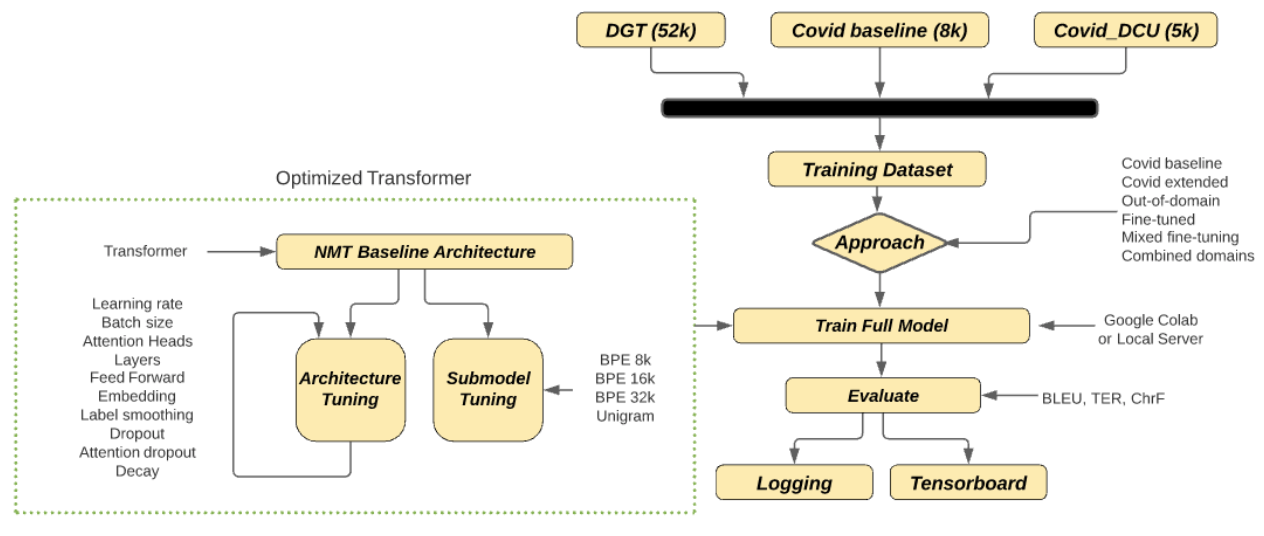}
    \caption{Proposed Approach. Optimal hyperparameters are applied to Transformer models which are trained using one of several possible approaches. The training dataset composition is determined by the chosen approach. Models are subsequently evaluated using a suite of metrics.}
    \label{fig:approach}
\end{figure}

Hyperparameter optimization of Recurrent Neural Network (RNN) models in low-resource settings has previously demonstrated considerable performance improvements~\citep{sennrich2019revisiting}. The extent to which such optimization techniques may be applied to Transformer models in similar low-resource scenarios was evaluated in a previous study~\citep{lankford2021Transformer}. Evaluations included modifying the number of attention heads, the number of layers and experimenting with regularization techniques such as dropout and label smoothing. Most importantly, the choice of subword model type and the vocabulary size were evaluated.

In order to test the effectiveness of our approach, models were trained using three English-Irish parallel datasets: a general corpus of 52k lines from the Directorate General for Translation (DGT) and two in-domain corpora of Covid data (8k and 5k lines). All experiments involved concatenating source and target corpora to create a shared vocabulary and a shared SentencePiece~\citep{kudo2018SentencePiece} subword model.  The impact of using separate source and target subword models was not explored.

The approach adopted is illustrated in Figure \ref{fig:approach} and the datasets used in evaluating this approach are outlined in Table \ref{tab:approach}. All models were developed using a Transformer architecture. 

\begin{table}
\centering
\begin{tabular}{llllll}
\hline
\textbf{Approach} &
 \textbf{Source} &
 \textbf{Lines} & \\ \\ \hline
Covid baseline & Baseline & 8k &  \\
Covid extended & Baseline + Covid\_DCU & 13k & \\
Out-of-domain & DGT & 52k &  \\
Fine-tuned & Baseline + Covid\_DCU + DGT & 65k & \\
Mixed fine-tuned & Baseline + Covid\_DCU + DGT & 65k & \\
Combined domains & Baseline + Covid\_DCU + DGT & 65k & \\ \hline
\end{tabular}
\caption{Datasets used in proposed approach}
\label{tab:approach}
\end{table}

\subsection{Architecture Tuning}

Long training times associated with NMT make it costly to tune systems using conventional Grid Search approaches. A previous study identified the hyperparameters required for optimal performance ~\citep{lankford2021Transformer}. Reducing the number of hidden layer neurons and increasing dropout led to significantly better performance.  Furthermore, within the context of low-resource English to Irish translation, using a 16k BPE submodel resulted in the highest performing models. The Transformer hyperparameters, chosen in line with these findings, are outlined in Table \ref{tab:hpo-table}.
\begin{center}
\begin{table}
\center
\begin{tabular}{ll}
\hline
\textbf{Hyperparameter} & \textbf{Values}                \\ \hline
Learning rate            & 0.1, 0.01, 0.001, \textbf{2}            \\ \hline
Batch size               & 1024, \textbf{2048},  4096, 8192       \\ \hline
Attention heads          & \textbf{2}, 4, \textbf{8}                     \\ \hline
Number of layers         & 5, \textbf{6}                           \\ \hline
Feed-forward dimension   & \textbf{2048}                           \\ \hline
Embedding dimension      & 128, \textbf{256}, 512                  \\ \hline
Label smoothing          & \textbf{0.1}, 0.3                       \\ \hline
Dropout                  & 0.1, \textbf{0.3}                       \\ \hline
Attention dropout        & \textbf{0.1}                            \\ \hline
Average Decay            & 0, \textbf{0.0001}                      \\ \hline
\end{tabular}
\caption{Hyperparameter optimization for Transformer models. Optimal parameters are highlighted in bold~\citep{lankford2021Transformer}.}
\label{tab:hpo-table}
\end{table}
\end{center}

\section{Empirical Evaluation}

\subsection{Experimental Setup}
\subsubsection{Datasets}
The performance of the Transformer approach is evaluated on English to Irish parallel datasets in the Covid domain. Three datasets were used in the evaluation of our models. These consisted of a baseline Covid dataset (8k) provided by MT Summit 2021~\citep{ojha-etal-2021-findings}, an in-domain Covid dataset (5k) developed at DCU and a publicly available out-of-domain dataset (52k) provided by DGT~\citep{steinberger2013dgt}. 

\subsubsection{Infrastructure}
Models were developed using a lab of machines each of which has an AMD Ryzen 7 2700X processor, 16 GB memory, a 256 SSD and an NVIDIA GeForce GTX 1080 Ti. Rapid prototype development was enabled through a Google Colab Pro subscription using NVIDIA Tesla P100 PCIe 16 GB graphic cards and up to 27GB of memory when available~\citep{bisong2019google}.

Our MT models were trained using the Pytorch implementation of OpenNMT 2.0, an open-source toolkit for NMT~\citep{klein2017opennmt}. 

\subsubsection{Metrics}

Automated metrics were used to determine the translation quality. All models were trained and evaluated using the BLEU~\citep{papineni2002BLEU}, TER~\citep{snover2006study} and ChrF~\citep{popovic2015ChrF} evaluation metrics. Case-insensitive BLEU scores, at the corpus level, are reported. Model training was stopped once an early stopping criteria of no improvement in validation accuracy for 4 consecutive iterations was recorded.

\subsection{Results}

Experimental results achieved using a Transformer architecture, with either 2 or 8 attention heads, are summarized in Table \ref{tab:trans-2heads} and in Table \ref{tab:trans-8heads}. Clearly in the context of our low-resource experiments, it can be seen there is little performance difference using Transformer architectures with a differing number of attention heads. The largest difference occurs when using a fine-tuned approach (2.1 BLEU points). However the difference between a 2 head and an 8 head approach is less than 1 BLEU point for all other models. The highest performing approach uses the extended Covid dataset (13k) which is a combination of the MT summit Covid baseline and a custom DCU Covid dataset. This Transformer model, with 2 heads, performs well across all key translation metrics (BLEU: 36.0, TER: 0.63 and ChrF3: 0.32).

\begin{figure}
     \begin{subfigure}[b]{0.5\textwidth}
         \centering
         \includegraphics[width=\textwidth]{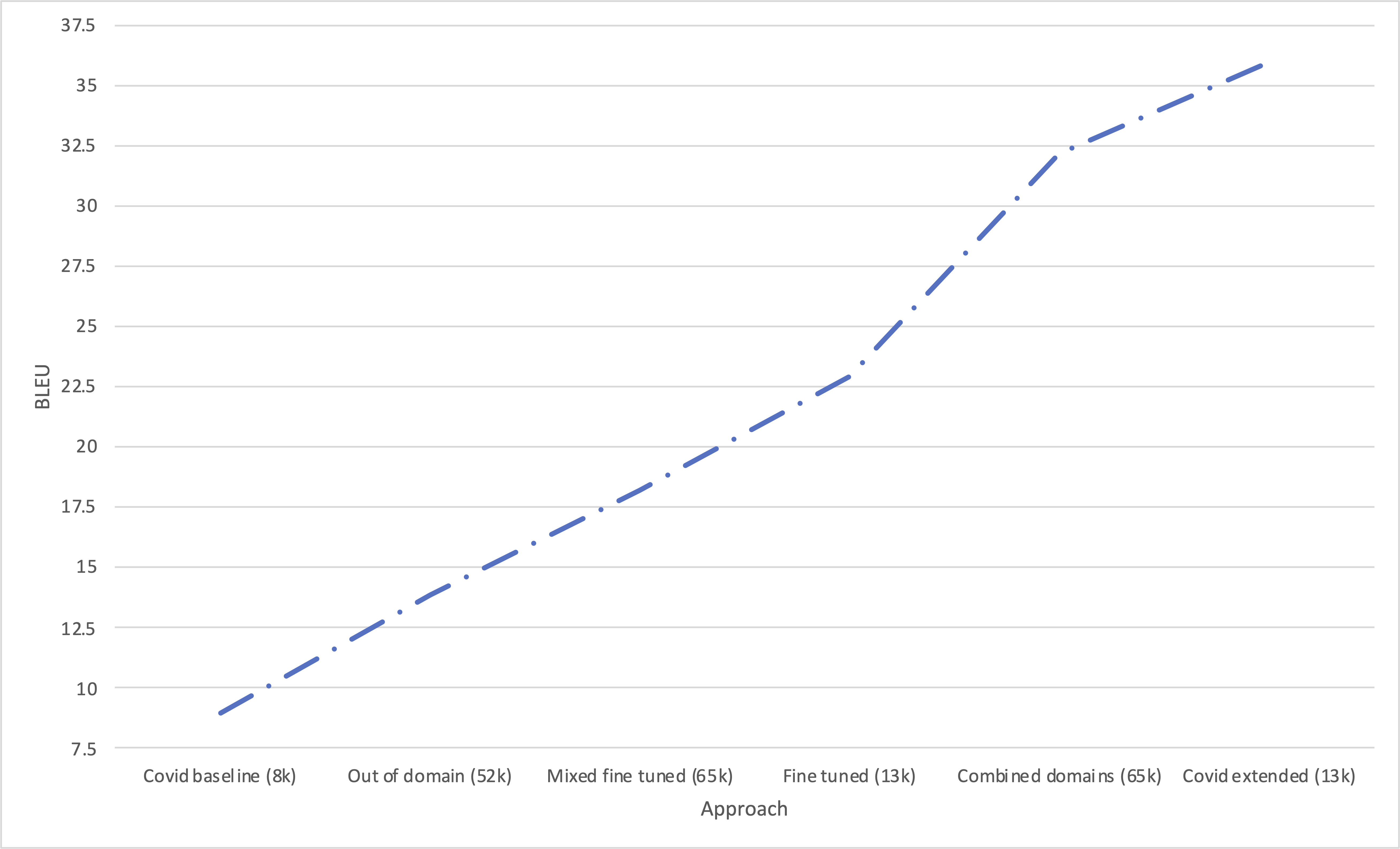}
         \caption{BLEU}
         \label{fig:bleu}
     \end{subfigure}
     \begin{subfigure}[b]{0.5\textwidth}
         \centering
         \includegraphics[width=\textwidth]{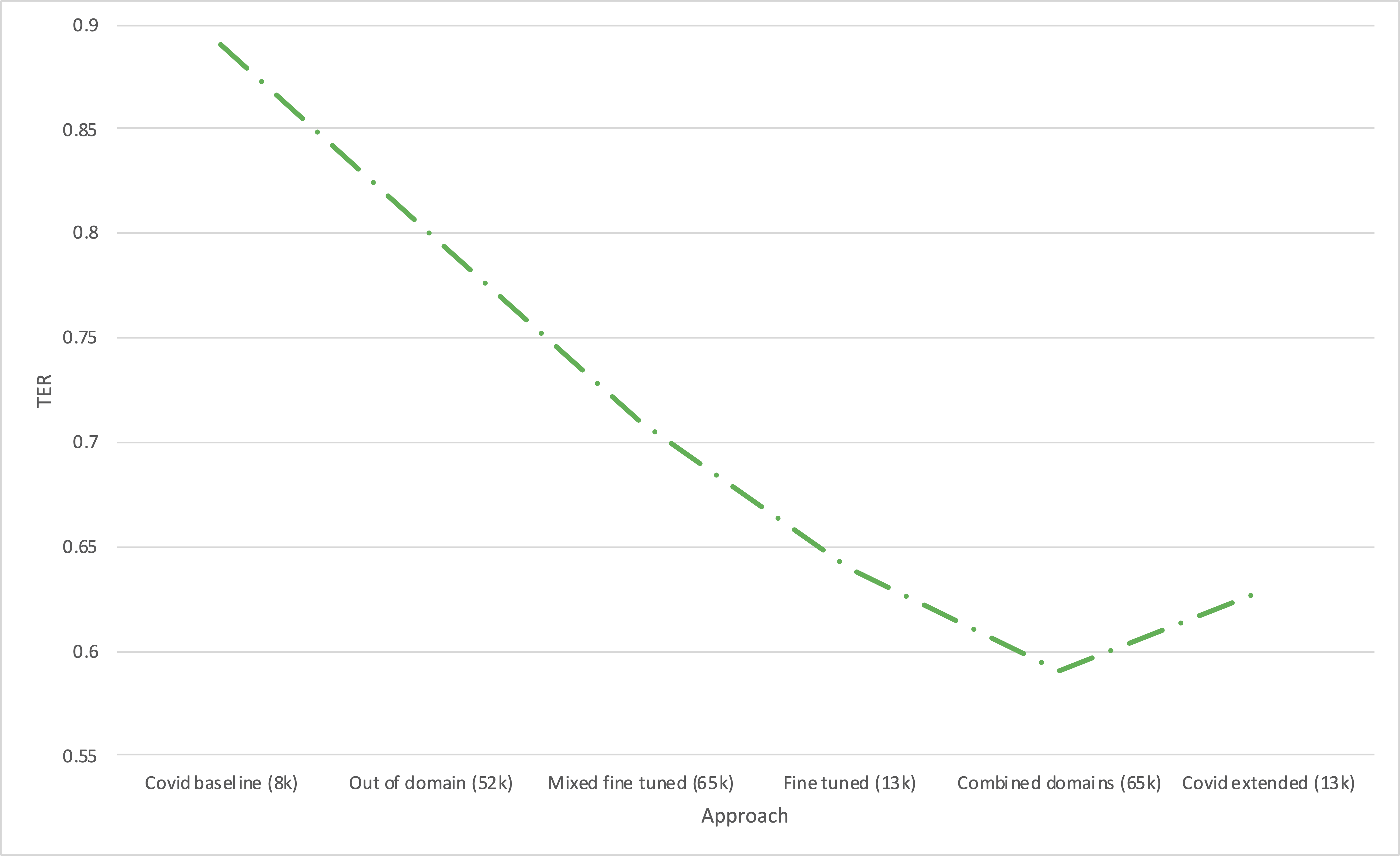}
         \caption{TER}
         \label{fig:ter}
     \end{subfigure}
        \caption{Translation performance of all approaches using Transformers with 2 heads}
        \label{fig:transper}
\end{figure}

\begin{table}[]
\centering
\begin{tabular}{lcccccc}
\hline
\textbf{System} & \multicolumn{1}{l}{\textbf{Heads}} & \multicolumn{1}{l}{\textbf{Lines}} & \multicolumn{1}{l}{\textbf{Steps}} & \multicolumn{1}{l}{\textbf{BLEU}$\uparrow$} & \multicolumn{1}{l}{\textbf{TER}$\downarrow$} & \multicolumn{1}{l}{\textbf{ChrF3}$\uparrow$} \\ \hline
Covid baseline & 2 & 8k & 35k & 9.0 & 0.89 & 0.32 \\ 
Covid extended & 2 & 13k & 35k & 36.0 & 0.63 & 0.54 \\ 
Out-of-domain & 2 & 52k & 200k & 13.9 & 0.80 & 0.41 \\ 
Fine-tuned & 2 & 65k & 35k & 22.9 & 0.64 & 0.42 \\ 
Mixed fine-tuned & 2 & 65k & 35k & 18.2 & 0.71 & 0.42 \\ 
Combined domains & 2 & 65k & 35k & 32.2 & 0.59 & 0.55 \\ \hline
\end{tabular}
\caption{Comparison of optimized Transformer performance with 2 attention heads}
\label{tab:trans-2heads}
\end{table}

\begin{table}[]
\centering
\begin{tabular}{lcccccc}
\hline
\textbf{System} & \multicolumn{1}{l}{\textbf{Heads}} & \multicolumn{1}{l}{\textbf{Lines}} & \multicolumn{1}{l}{\textbf{Steps}} & \multicolumn{1}{l}{\textbf{BLEU}$\uparrow$} & \multicolumn{1}{l}{\textbf{TER}$\downarrow$} & \multicolumn{1}{l}{\textbf{ChrF3}$\uparrow$} \\ \hline
Covid baseline & 8 & 8k & 35k & 9.6 & 0.91 & 0.33 \\ 
Covid extended & 8 & 13k & 35k & 35.7 & 0.61 & 0.55 \\ 
Out-of-domain & 8 & 52k & 200k & 13.0 & 0.80 & 0.40 \\ 
Fine-tuned & 8 & 65k & 35k & 25.0 & 0.63 & 0.43 \\ 
Mixed fine-tuned & 8 & 65k & 35k & 18.0 & 0.71 & 0.42 \\ 
Combined domains & 8 & 65k & 35k & 32.8 & 0.59 & 0.57 \\ \hline
\end{tabular}
\caption{Comparison of optimized Transformer performance with 8 attention heads}
\label{tab:trans-8heads}
\end{table}

The worst performing model uses the Covid baseline which is not surprising given that only 8k lines are available. The performance of the higher resourced models (out-of-domain, fine-tuned, mixed fine-tuned and combined domains) all lag that of the Covid extended model. In particular, the out-of-domain model, using the DGT dataset, performs very poorly with a BLEU score of just 13.9 on a Transformer model with 2 heads.

The BLEU and TER scores for all approaches are illustrated in Figure \ref{fig:bleu} and Figure \ref{fig:ter}. As expected, there is a high level of inverse correlation between BLEU and TER. Well-performing models, with high BLEU scores, also required little post editing effort as indicated by their lower TER scores. 

\section{Discussion}

Standard Transformer parameters identified in a previous study were observed to perform well~\citep{lankford2021Transformer}. Reducing hidden neurons to 256 and increasing regularization dropout to 0.3 improved translation performance and these hyperparameters were chosen when building all Transformer models. Furthermore a batch size of 2048 and using 6 layers for the encoder / decoder were chosen throughout. 

The results demonstrate that translation performance for specific domains is driven by the amount of data which is available for that specific domain. It is noteworthy that an in-domain dataset of 13k lines (Covid extended), trained for just 35k steps outperformed by 22.1 BLEU points the corresponding out-of-domain 52k dataset (DGT) which was trained for 200k steps. 

\section{Conclusion and Future Work}
In the official evaluation for LoResMT 2021, our English–Irish system was ranked first according to the BLEU, TER and ChrF scores. We demonstrate that a high performing in-domain translation model can be built with a dataset of 13k lines. Developing a small in-domain dataset, of just 5k lines, improved the BLEU score by 27 points when models were trained with the combined Covid baseline and custom Covid dataset.     

Following on from our previous work, careful selection of Transformer hyperparameters, and using a 16k BPE SentencePiece submodel, enabled rapid development of high performing translation models in a low-resource setting. 

Within the context of our research in low-resource English to Irish translation, we have shown that augmenting in-domain data, by a small amount, performed better than approaches which incorporate fine-tuning, mixed fine-tuning or the combination of domains.    

As part of our future work, we plan to develop English-Irish MT models trained on a dataset derived from the health domain. Domain adaptation, through fine-tuning such models with the Covid extended dataset may further improve Covid MT performance.     

\section*{Acknowledgements}
This work was supported by ADAPT,   which is funded under the SFI Research Centres Programme (Grant 13/RC/2016) and is co-funded by the European Regional Development Fund. This research was also funded by the Munster Technological University.
\small

\bibliographystyle{apalike}
\bibliography{mtsummit2021}

\begin{thebibliography}{}

\bibitem[Araabi and Monz, 2020]{araabi2020optimizing}
Araabi, A. and Monz, C. (2020).
\newblock Optimizing transformer for low-resource neural machine translation.
\newblock {\em arXiv preprint arXiv:2011.02266}.

\bibitem[Bisong, 2019]{bisong2019google}
Bisong, E. (2019).
\newblock Google colaboratory.
\newblock In {\em Building Machine Learning and Deep Learning Models on Google Cloud Platform}, pages 59--64. Springer.

\bibitem[Chu et~al., 2017]{chu2017empirical}
Chu, C., Dabre, R., and Kurohashi, S. (2017).
\newblock An empirical comparison of domain adaptation methods for neural machine translation.
\newblock In {\em Proceedings of the 55th Annual Meeting of the Association for Computational Linguistics (Volume 2: Short Papers)}, pages 385--391.

\bibitem[Crego et~al., 2016]{crego2016systran}
Crego, J., Kim, J., Klein, G., Rebollo, A., Yang, K., Senellart, J., Akhanov, E., Brunelle, P., Coquard, A., Deng, Y., et~al. (2016).
\newblock Systran's pure neural machine translation systems.
\newblock {\em arXiv preprint arXiv:1610.05540}.

\bibitem[Klein et~al., 2017]{klein2017opennmt}
Klein, G., Kim, Y., Deng, Y., Senellart, J., and Rush, A.~M. (2017).
\newblock Opennmt: Open-source toolkit for neural machine translation.
\newblock {\em arXiv preprint arXiv:1701.02810}.

\bibitem[Kudo and Richardson, 2018]{kudo2018SentencePiece}
Kudo, T. and Richardson, J. (2018).
\newblock Sentencepiece: A simple and language independent subword tokenizer and detokenizer for neural text processing.
\newblock {\em arXiv preprint arXiv:1808.06226}.

\bibitem[Lankford et~al., 2021]{lankford2021Transformer}
Lankford, S., Alfi, H., and Way, A. (2021).
\newblock Transformers for low resource languages: Is feidir linn.
\newblock In {\em Proceedings of the 18th Conference of the Association for Machine Translation in the Americas (Volume 1: Research Papers)}.

\bibitem[Ojha et~al., 2021]{ojha-etal-2021-findings}
Ojha, A.~K., Liu, C.-H., Kann, K., Ortega, J., Satam, S., and Fransen, T. (2021).
\newblock Findings of the {LoResMT 2021 Shared Task on COVID and Sign Language for Low-Resource Languages}.
\newblock In {\em Proceedings of the 4th Workshop on Technologies for MT of Low Resource Languages}.

\bibitem[Papineni et~al., 2002]{papineni2002BLEU}
Papineni, K., Roukos, S., Ward, T., and Zhu, W.-J. (2002).
\newblock Bleu: a method for automatic evaluation of machine translation.
\newblock In {\em Proceedings of the 40th annual meeting of the Association for Computational Linguistics}, pages 311--318.

\bibitem[Popovi{\'c}, 2015]{popovic2015ChrF}
Popovi{\'c}, M. (2015).
\newblock chrf: character n-gram f-score for automatic mt evaluation.
\newblock In {\em Proceedings of the Tenth Workshop on Statistical Machine Translation}, pages 392--395.

\bibitem[Sennrich and Zhang, 2019]{sennrich2019revisiting}
Sennrich, R. and Zhang, B. (2019).
\newblock Revisiting low-resource neural machine translation: A case study.
\newblock {\em arXiv preprint arXiv:1905.11901}.

\bibitem[Snover et~al., 2006]{snover2006study}
Snover, M., Dorr, B., Schwartz, R., Micciulla, L., and Makhoul, J. (2006).
\newblock A study of translation edit rate with targeted human annotation.
\newblock In {\em Proceedings of association for machine translation in the Americas}, volume 200. Citeseer.

\bibitem[Steinberger et~al., 2013]{steinberger2013dgt}
Steinberger, R., Eisele, A., Klocek, S., Pilos, S., and Schl{\"u}ter, P. (2013).
\newblock Dgt-tm: A freely available translation memory in 22 languages.
\newblock {\em arXiv preprint arXiv:1309.5226}.

\bibitem[Vaswani et~al., 2017]{vaswani2017attention}
Vaswani, A., Shazeer, N., Parmar, N., Uszkoreit, J., Jones, L., Gomez, A.~N., Kaiser, L., and Polosukhin, I. (2017).
\newblock Attention is all you need.
\newblock {\em arXiv preprint arXiv:1706.03762}.

\bibitem[Wu et~al., 2016]{wu2016google}
Wu, Y., Schuster, M., Chen, Z., Le, Q.~V., Norouzi, M., Macherey, W., Krikun, M., Cao, Y., Gao, Q., Macherey, K., et~al. (2016).
\newblock Google's neural machine translation system: Bridging the gap between human and machine translation.
\newblock {\em arXiv preprint arXiv:1609.08144}.

\end{thebibliography}

\end{document}